\documentclass[10pt,twocolumn,letterpaper]{article}

\usepackage[pagenumbers]{cvpr} %

\usepackage[dvipsnames]{xcolor}

\newcommand{\boldparagraph}[1]{\vspace{0.2cm}\noindent{\bf #1:} }
\definecolor{cvprblue}{rgb}{0.21,0.49,0.74}
\usepackage[pagebackref,breaklinks,colorlinks,citecolor=cvprblue]{hyperref}
\usepackage{comment}
\usepackage{xfrac}
\usepackage{arydshln}
\def\confName{CVPR}
\def\confYear{2024}

\title{Morphable Diffusion: \\3D-Consistent Diffusion for Single-image Avatar Creation}

\author{Xiyi Chen$^1$ \quad Marko Mihajlovic$^1$  \quad Shaofei Wang$^{1,2,3}$ \quad Sergey Prokudin$^{1,4}$ \quad Siyu Tang$^1$\\
\\ETH Zürich$^1$; University of T\"{u}bingen$^2$; T\"{u}bingen AI Center$^3$;\\ ROCS, University Hospital Balgrist, University of Zürich$^4$\\[4pt]
{\url{https://xiyichen.github.io/morphablediffusion/}}
}

\begin{document}

\twocolumn[{%
    \renewcommand\twocolumn[1][]{#1}%
    \setlength{\tabcolsep}{0.0mm} %

    \newcommand{\sz}{0.125}  %

    \maketitle
    \begin{center}
        \newcommand{\teaserwidth}{\textwidth}
    \vspace{-0.3in}
        \includegraphics[width=\linewidth]{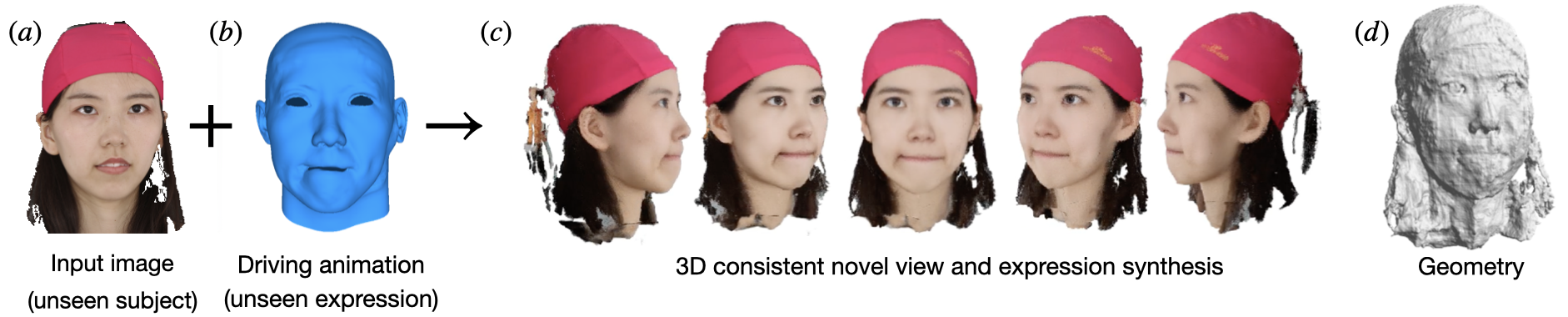}
      \vspace{-0.3cm}
        \captionof{figure}{\textbf{Morphable diffusion}. We introduce a morphable diffusion model to enable consistent controllable novel view synthesis of humans from a single image. 
        Given a single input image \textit{(a)} and a morphable mesh model with a target facial expression \textit{(b)} our method directly generates 3D consistent and photo-realistic images from novel viewpoints \textit{(c)}. Using the generated multi-view consistent images, we can reconstruct a coarse 3D model \textit{(d)} using off-the-shelf neural surface reconstruction methods such as \cite{neus2}.  
        } 

    \label{fig:teaser}
    
    \end{center}%
}]

\begin{abstract}
Recent advances in generative diffusion models have enabled the previously unfeasible capability of generating 3D assets from a single input image or a text prompt. In this work, we aim to enhance the quality and functionality of these models for the task of creating controllable, photorealistic human avatars. We achieve this by integrating a 3D morphable model into the state-of-the-art multi-view-consistent diffusion approach. We demonstrate that accurate conditioning of a generative pipeline on the articulated 3D model enhances the baseline model performance on the task of novel view synthesis from a single image. More importantly, this integration facilitates a seamless and accurate incorporation of facial expression and body pose control into the generation process. To the best of our knowledge, our proposed framework is the first diffusion model to enable the creation of fully 3D-consistent, animatable, and photorealistic human avatars from a single image of an unseen subject; extensive quantitative and qualitative evaluations demonstrate the advantages of our approach over existing state-of-the-art avatar creation models on both novel view and novel expression synthesis tasks. The code for our project is publicly available. 
\end{abstract}
    
\section{Introduction}
\label{sec:intro}

The field of photorealistic controllable human avatar generation has been subject to several technological leaps in the recent decade. 
The introduction of large 3D scan collections has facilitated the construction of expressive, articulated models of 3D human bodies \cite{loper2015smpl,SMPL-X:2019}, faces \cite{blanz2023morphable,paysan20093d,FLAME:SiggraphAsia2017}, and hands \cite{MANO:SIGGRAPHASIA:2017}. From the outset, one of the primary applications of these models was to reconstruct a 3D avatar from highly under-constrained setups, such as monocular video or a single image~\cite{blanz2023morphable,SMPL-X:2019,Feng:SIGGRAPH:2021,kocabas2019vibe}. 

While allowing for rich semantic information to be inferred, these 3D morphable models were limited in the level of photorealism due to their focus on minimally clothed bodies and face regions, as well as their reliance on the standard mesh-based computer graphics pipelines for rendering.

Recently, the task of generating \textit{photorealistic} avatars \cite{peng2021neural} gained significant attention from the research community due to its potential to revolutionize our ways of digital communication. Here, combining novel neural rendering techniques \cite{mildenhall2020nerf,xie2022neural} with articulated human models allowed for a new level of generated image quality. However, the best-performing models here still require a significant visual input, such as calibrated multi-view images \cite{habermann2021,wang2022arah} or monocular video sequences of the subject \cite{zheng2022avatar,zielonka2023instant,weng_humannerf_2022_cvpr,qian20233dgsavatar}. 

Concurrently, the field of generative modeling emerged with its ability to create highly photorealistic assets by learning a complex distribution of real-world image data \cite{karras2019style,rombach2022high}. Here, a substantial body of research has been dedicated to better controlling the generative process of such models in the case of human-related imagery \cite{dong2023ag3d,zhang2023adding,isola2017image}. This is regularly done by conditioning the pipelines on 2D keypoints \cite{chan2019everybody}, 3D deformable models \cite{tewari2020stylerig}, or text \cite{pan2023avatarstudio}. Another emerging topic is building geometry-aware generative models that allow view-consistent image synthesis and 3D asset extraction \cite{chan2022efficient,poole2022dreamfusion,liu2023syncdreamer}.

The method proposed in this work can be considered the next logical step in the evolution and unification of the three aforementioned branches of research on photorealistic controllable virtual humans. We begin with investigating the performance of a state-of-the-art view-consistent diffusion model \cite{Liu_2023_ICCV,liu2023syncdreamer} on the task of novel view synthesis of human heads and full bodies. 
However, we show that a simple finetuning of the baseline on a limited dataset (e.g.~\cite{yang2020facescape,tao2021function4d}) leads to sub-optimal results, with the model failing to preserve the identity or the facial expression / body pose.

We follow in the footsteps of the view-consistent diffusion model and propose a novel finetuning strategy and architecture that enhances the reconstruction quality and allows animation. Our key idea is to leverage a well-studied statistical model of human shapes \cite{FLAME:SiggraphAsia2017, loper2015smpl} to introduce human prior and guide the reconstruction process. More importantly, a controllable 3D model also allows us to perform a more challenging task of photorealistic animation of a human head from a single image under novel expression (Figure~\ref{fig:teaser}). 

A similar type of diffusion process guidance has been investigated by several prior works \cite{ding2023diffusionrig,zhang2023adding}. However, as we will show later, the models conditioned purely on 2D rasterizations of meshes or projected keypoints struggle to achieve truly 3D-consistent novel view and expression synthesis results. 

\begin{figure*}[h!]
\centering
\includegraphics[width=\textwidth]{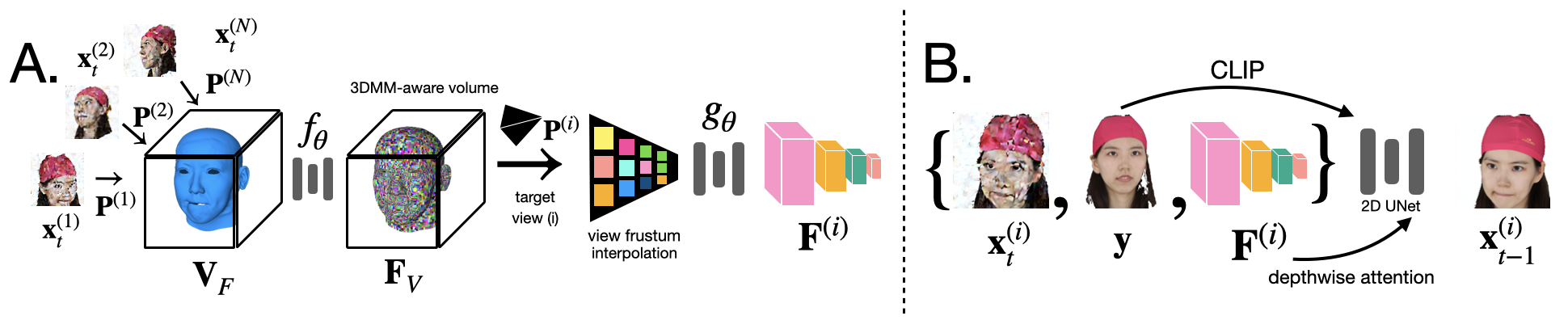}
\vspace{-0.5cm}
\caption{\textbf{Morphable diffusion step.} This figure gives an overview of a single denoising step of the proposed 3D morphable diffusion pipeline. Our morphable denoiser takes as input a single image $\mathbf{y}$ and the underlying human model $\mathcal{M}$ and generates $N$ novel views from pre-defined viewpoints. Given the noisy images of N fixed views $x_t^{(1:N)}$ obtained from the previous iteration, camera projection matrices $P^{(1:N)} = (K^{(1:N)}, R^{(1:N)}, T^{(1:N)})$, and the target articulated 3D mesh model, we construct %
{\it A)} a morphable noise volume by attaching the 2D noise features onto mesh vertices that are processed by a SparseConvNet $f_\theta$ to output a 3DMM-aware feature volume $\mathbf{F}_V$, which is further interpolated to the frustum $\mathbf{F}^{(i)}$ of a target view $(i)$ that we wish to synthesize. 
{\it B)} The noisy target image $\mathbf{x}_t^{(i)}$, the input image $\mathbf{y}$, and the target feature frustum are then processed by a pre-trained 2D UNet akin to \cite{liu2023syncdreamer} to predict the denoised image in the next iteration $\mathbf{x}_{t-1}^{(i)}$.
}
\label{fig:overview}
\end{figure*}

To alleviate this problem, we introduce the conditioning of the diffusion process on a 3D morphable model that performs an uplifting of the noisy image features and associates them with the corresponding mesh vertices \textit{in 3D space} (Figure~\ref{fig:overview}A). 
We also introduce a shuffled training scheme for the diffusion model: during training, the model is trained to predict a view-consistent image set of novel facial expressions, given a single head image with a different expression as a reference and an articulated 3D model as a driving signal. 
The resulting combination of a powerful diffusion network and 3D model conditioning allows for the first-time building of a highly photorealistic animatable head model of an unseen subject from a single input image with an unseen facial expression as driving signal.

To summarize, our contributions are as follows: (a) We analyze the applicability of state-of-the-art multi-view consistent diffusion models for the task of human avatar creation and propose a superior pipeline that consistently improves the quality of generated images across most metrics, thanks to the efficient conditioning of the generative process on a deformable 3D model; (b) We further propose a more efficient training scheme to enable the generation of new facial expressions for an unseen subject from a single image.

\section{Related work} \label{sec:related} 

\boldparagraph{3D morphable models} %
Traditional methods for constructing shapes have been predominantly centered around parameterized 3D mesh representations, particularly in the context of human faces \cite{blanz2023morphable,paysan20093d,FLAME:SiggraphAsia2017}, bodies \cite{loper2015smpl, SMPL-X:2019, xu2020ghum}, and hands~\cite{MANO:SIGGRAPHASIA:2017,li2022nimble}. 
These data-driven models are typically created from high-quality 3D scans obtained in expensive studio environments and they have laid the foundation for reconstructing humans in the wild from as few as a single input image \cite{kanazawa2018end,kocabas2019vibe,Feng:SIGGRAPH:2021}. 
However, these mesh-based representations are limited in quality for representing thin geometric details (\eg hair) and are not suitable for truly photo-realistic novel view synthesis.

\boldparagraph{Photorealistic avatars} %
Neural radiance fields (NeRFs) \cite{mildenhall2020nerf} have recently enabled rapid development of modeling and reconstructing photorealistic scenes. 
Several recent works have successfully combined neural rendering with articulated human models \cite{peng2021neural, xie2022neural, Lombardi21} to enable high-quality photorealistic avatar reconstruction. 
However, they still require a substantial visual input such as a monocular video \cite{zheng2022avatar,zielonka2023instant,weng_humannerf_2022_cvpr,jiang2023instantavatar,qian20233dgsavatar} or several calibrated multi-view images or videos \cite{habermann2021,wang2022arah, mihajlovic2022keypointnerf, chen2023gm, kania2022conerf}. 

\boldparagraph{Avatars from a single image} %
Single-image human avatar creation has been an active research area in the past few years. 
The seminal work PIFu~\cite{saito2019pifu} has proposed to directly model the human body shape as an implicit representation conditioned on pixel-aligned features. 
Several follow-ups \cite{saito2020pifuhd, alldieck2022photorealistic, he2020geoPifu, huang2020arch, he2021archPlus, xiu2023econ, xiu2022icon} have focused on improving robustness and reconstruction quality; others have focused on reconstructing human faces \cite{rebain2022lolnerf,burkov2023multi,trevithick2023real,khakhulin2022realistic,chu2024gpavatar}.
However, these methods lack controllability and cannot be directly animated or used to synthesize novel views for diverse novel expressions or poses of an unseen subject.

\boldparagraph{Controllable face avatars} Controllable face avatars allow rigging of appearance, identity, pose, or expression from a single image \cite{zhuang2022mofanerf, hong2021headnerf, li2023goha}. However, these methods only demonstrate the manipulation ability under frontal views and produce incomplete or blurry renderings under extreme head poses, as we will demonstrate in the following sections. Several other works achieve the rigging from multiple views or monocular videos of the input subject \cite{zheng2022avatar, Zheng2023pointavatar, zielonka2023instant}. Closer to our work is DiffusionRig~\cite{ding2023diffusionrig} that leverages the photorealism of the diffusion models \cite{rombach2022high} with the explicit controllability achieved via conditioning of the denoising process on 2D rasterizations of a morphable model. 
However, this work struggles to achieve a multi-view consistent novel view synthesis, as we will demonstrate in the experiment section later.

\boldparagraph{Generative models} %
As reconstruction from a single view is inherently a challenging ill-posed task, many works have attempted to formulate the problem as a generative task instead. 
A rich set of related methods leverage generative adversarial networks \cite{karras2019style} or powerful diffusion models \cite{rombach2022high} to learn complex data distributions from real-world imagery and to directly synthesize novel images without modeling an underlying 3D model. 
In the context of human avatars \cite{dong2023ag3d,zhang2023adding,isola2017image}, prior works explore different input conditions to control the generative process such as keypoints~\cite{chan2019everybody}, deformable models~\cite{tewari2020stylerig}, or just textual queries~\cite{kolotouros2023dreamhuman,cao2023dreamavatar,pan2023avatarstudio}. 
However, achieving a high degree of photorealism in these models comes at the expense of 3D consistency.

\boldparagraph{3D-consistent generative models}
Several attempts have been made towards constructing a 3D consistent generative model \cite{chan2022efficient, bergman2022generative,sun2023next3d,shi2022deep,or2022stylesdf,nguyen2020blockgan,nguyen2019hologan,liao2020towards,wu2016learning,wu20233dportraitgan,wang2023rodin,Schwarz2022NEURIPS,gu2021stylenerf,An_2023_CVPR}. 
On the other side, the recent work Zero-1-to-3~\cite{Liu_2023_ICCV} learns to control the camera perspective in large-scale diffusion models and enable consistent generation of novel views. 
SyncDreamer~\cite{liu2023syncdreamer} further improves the multi-view consistency through a 3D-aware feature attention module. 
However, the novel view synthesis from a single image of this model cannot be explicitly controlled, \eg by a human expression or pose, which is essential for avatar creation. 
In this work, we build on the state-of-the-art multi-view consistent diffusion framework~\cite{liu2023syncdreamer} and enhance the reconstruction quality while enabling additional control for generating animatable photorealistic avatars.

\boldparagraph{Finetuning diffusion models} 
The main limitation of diffusion models is the long and expensive training. 
Therefore, many recent works \cite{hu2021lora, dettmers2023qlora, liu2023syncdreamer, liu2023one2345} focus instead on finetuning large pre-trained models. 
Here, a popular approach~\cite{zhang2023adding} is to inject the control parameters as an additional signal when finetuning. 
However, adjusting this framework operating in 2D space to finetune multi-view consistent diffusion models \cite{liu2023syncdreamer} which contains a 3D conditioning module is not straightforward. In the following, we will examine how to effectively control the diffusion models for photorealistic avatar creation. 

\section{Morphable diffusion model}
Given an input image $\mathbf{y}$ of a human, our goal is to synthesize $N$ multi-view consistent novel views from predefined viewing angles and allow for their animation via a morphable model $\mathcal{M}$. 
Our method builds on the recent multi-view consistent diffusion model \cite{liu2023syncdreamer} while further improving the reconstruction quality and allowing for explicit manipulation of the synthesized images. 
We start by reviewing the 3D-consistent diffusion model (Sec.~\ref{subsec:Preliminaries}) and later detail our morphable diffusion model (Sec.~\ref{subsec:MorphableDiffusion}).

\subsection{Preliminaries: multi-view diffusion} \label{subsec:Preliminaries}
Multi-view diffusion \cite{liu2023syncdreamer} consistently learns the joint distribution of novel views at $N$ fixed viewpoints $\{\mathbf{x}_0^{(i)}\}_{i=1}^N$ given an input condition $\textbf{y}$: $p_\theta(\mathbf{x}_0^1, \dots, \mathbf{x}_0^N,|\mathbf{y})$; $\mathbf{y}$ is omitted for brevity in the following. 
Analogously to the diffusion model \cite{sohl2015deep}, it defines the \textit{forward process} 
\begin{equation}
\label{eq:mv_forward}
    q(\mathbf{x}^{(1:N)}_{1:T}|\mathbf{x}_0^{(1:N)}) = \prod_{t=1}^T \prod_{n=1}^N q(\mathbf{x}^{(n)}_t|\mathbf{x}^{(n)}_{t-1}),
\end{equation}
\begin{equation}
    q(\mathbf{x}^{(n)}_t|\mathbf{x}^{(n)}_{t-1})=\mathcal{N}(\mathbf{x}^{(n)}_t;\sqrt{1-\beta_t} \mathbf{x}^{(n)}_{t-1},\beta_t \mathbf{I}),
\end{equation}
and the \textit{reverse process}
\begin{equation} 
\label{eq:mv_reverse}
    p_\theta(\mathbf{x}_{0:T}^{(1:N)}) = p(\mathbf{x}^{(1:N)}_T) \prod_{t=1}^{T} \prod_{n=1}^{N} p_\theta(\mathbf{x}^{(n)}_{t-1}|\mathbf{x}^{(1:N)}_{t}),
\end{equation}
\begin{equation}
    p_\theta(\mathbf{x}^{(n)}_{t-1}|\mathbf{x}^{(1:N)}_t)=\mathcal{N}(\mathbf{x}^{(n)}_{t-1};\mathbf{\mu}^{(n)}_\theta(\mathbf{x}^{(1:N)}_t,t),\sigma^2_t \mathbf{I}),
\end{equation}
where $\mathbf{\mu}^{(n)}_\theta$ is trainable, while $\beta_t$ and $\sigma_t$ are fixed time-dependent coefficients. 

To learn the joint distribution by minimizing the negative log-likelihood, multi-view diffusion \cite{liu2023syncdreamer} follows the DDPM's~\cite{ho2020denoising} parameterization:  
\begin{equation}
    \mathbf{\mu}^{(n)}_\theta(\mathbf{x}^{(1:N)}_t,t)=\frac{1}{\sqrt{\alpha}_t}\left(\mathbf{x}^{(n)}_t - \frac{\beta_t}{\sqrt{1-\bar{\alpha}_t}} \mathbf{\epsilon}^{(n)}_\theta (\mathbf{x}^{(1:N)}_t, t)\right),
\end{equation}
where $\epsilon_\theta$ is a trainable \textit{noise predictor} (in practice, parameterized by a UNet~\cite{RFB15a}) and other constants ($\alpha_t, \overline{\alpha}_t)$ are derived from $\beta_t$. 
Then, the training is performed by minimizing the loss
\begin{equation}
    \ell = \mathbb{E}_{t,\mathbf{x}^{(1:N)}_0,n,\mathbf{\epsilon}^{(1:N)}} \left[\|\mathbf{\epsilon}^{(n)} - \mathbf{\epsilon}^{(n)}_\theta (\mathbf{x}^{(1:N)}_t,t)\|_2\right],
\end{equation}
where $\mathbf{\epsilon}^{(1:N)} \sim \mathcal{N}(0, \mathbf{I})$ is the sampled Gaussian noise.

\subsection{Morphable multi-view diffusion model} \label{subsec:MorphableDiffusion}
To increase the reconstruction fidelity and extend the baseline model to animatable avatar, we introduce a morphable diffusion model for animatable high-fidelity multi-view consistent novel view synthesis of humans from a single input image. 

Our model takes as input a single 
image $\mathbf{y}$
and an underlying morphable model $\mathcal{M}$ (\eg SMPL~\cite{loper2015smpl}, FLAME~\cite{FLAME:SiggraphAsia2017}, or bilinear~\cite{yang2020facescape} model) that maps a low-dimension identity $\beta \in \mathbb{R}^{I}$ and expression $\Theta \in \mathbb{R}^{E}$ parameters to a mesh containing $n_v$ vertices:
\begin{equation}
    \mathcal{M}: \beta,\Theta \mapsto \mathbb{R}^{n_v \times 3}.
\end{equation}

Given the image $\mathbf{y}$ and mesh $\mathcal{M}$, we synthesize $N$ images $\mathbf{x}_0^{(1:N)}$ of the same resolution through an iterative diffusion process. 
These novel views are generated from fixed viewpoints with the relative camera translations $T^{(1:N)} \in \mathbb{R}^{3}$, rotations $R^{(1:N)} \in \mathbb{R}^{3 \times 3}$ and their respective calibration parameters $K^{(1:N)} \in \mathbb{R}^{3 \times 3}$. An overview of a single diffusion step is illustrated in Fig.~\ref{fig:overview}.

We leverage the mesh of the underlying morphable model to unproject and interpolate the $d$-dimensional target image noise features $x^{(1:N)}_0$ onto $n_v$ mesh vertices in the world space. These pixel-aligned noise features for all target views are then fused via channel-wise mean pooling to make the vertex features invariant of the order of target views. 
The output vertex features $\mathbf{V}_F \in \mathbb{R}^{n_v \times d}$ are then processed via a sparse 3D ConvNet~\cite{graham20183d, peng2021neural} $f_\theta$ and trilinearly interpolated to create a 3D morphable-model-aware feature volume:
\begin{equation}
    \mathbf{F}_V = f_\theta(\mathbf{V}_F) \in \mathbb{R}^{x \times y \times z \times f_V},
\end{equation}
where $x$, $y$, $z$, and $f_V$ denote the dimensionality and number of channels of the interpolated 3DMM-aware feature volume $\textbf{F}_{V}$.

Given a target view $i$, we uniformly sample $d_F$ points on every ray inside the view frustum defined by fixed near and far planes. This formulates a 3D grid $\mathbf{F}^{(i)} \in \mathbb{R}^{d_F \times h_F \times w_F \times f_V}$ inside the view frustum. We then assign the trilinearly interpolated features from $\mathbf{F}_{V}$ to each point in $\mathbf{F}^{(i)}$, which is then processed via a convolutional network to extract a feature volume $\mathbf{F}^{(i)}_{j}$ at each layer $j$ of the $L$-layer denoising UNet. This interpolation and the neural network operation are jointly denoted as $g_\theta$ and output a view frustum volume with $L$ grids:
\begin{equation}
    \mathbf{F}^{(i)}_{1:L} =  g_\theta(\mathbf{F}^{(i)}) = g_\theta(\mathbf{F}_{V} | R^{(i)},T^{(i)},K^{(i)}). 
\end{equation}

Lastly, the target $x_t^{(i)}$ and the CLIP encoding~\cite{radford2021learning} of the input image $\mathbf{y}$ are propagated through a pre-trained UNet that is conditioned on $\mathbf{F}^{(i)}_{1:L}$ through depth-wise attentions as in SyncDreamer. Specifically, a cross-attention \cite{vaswani2017attention} layer is applied on every intermediate feature map $f_j$ of the current view $i$ in the UNet:
\begin{equation}
    \text{Attention}(Q_j,K_j,V_j) = \text{softmax}(\frac{Q_jK_j^T}{\sqrt{d}}) \cdot V_j,
\end{equation}
with $Q_j = W_{Q, j} \cdot \varphi_j, K_j = W_{K, j} \cdot \mathbf{F}^{(i)}_{j}, V_j = W_{V, j} \cdot \mathbf{F}^{(i)}_{j}$, where $\varphi_j \in \mathbb{R}^{N \times d^{j}_{\epsilon} \times h_j \times w_j}$ is the intermediate representation of $f_j$ cross-attentioned with the CLIP feature of the input image, $\mathbf{F}^{(i)}_{j} \in {\mathbb{R}^{N \times d_r \times d_j \times h_j \times w_j}}$ is the feature volume downsampled from $\mathbf{F}^{(i)}$ corresponding to the $j$-th layer of the UNet. $W_{Q,j} \in \mathbb{R}^{d \times d^{j}_{\epsilon}}, W_{K, j} \in \mathbb{R}^{d \times d_r}, W_{V, j} \in \mathbb{R}^{d \times d_r}$ are learnable projection matrices \cite{rombach2022high}. The attention computation and feature aggregation is only performed along the depth dimension $d_j$.

\begin{figure*}
\centering
\includegraphics[width=\textwidth]{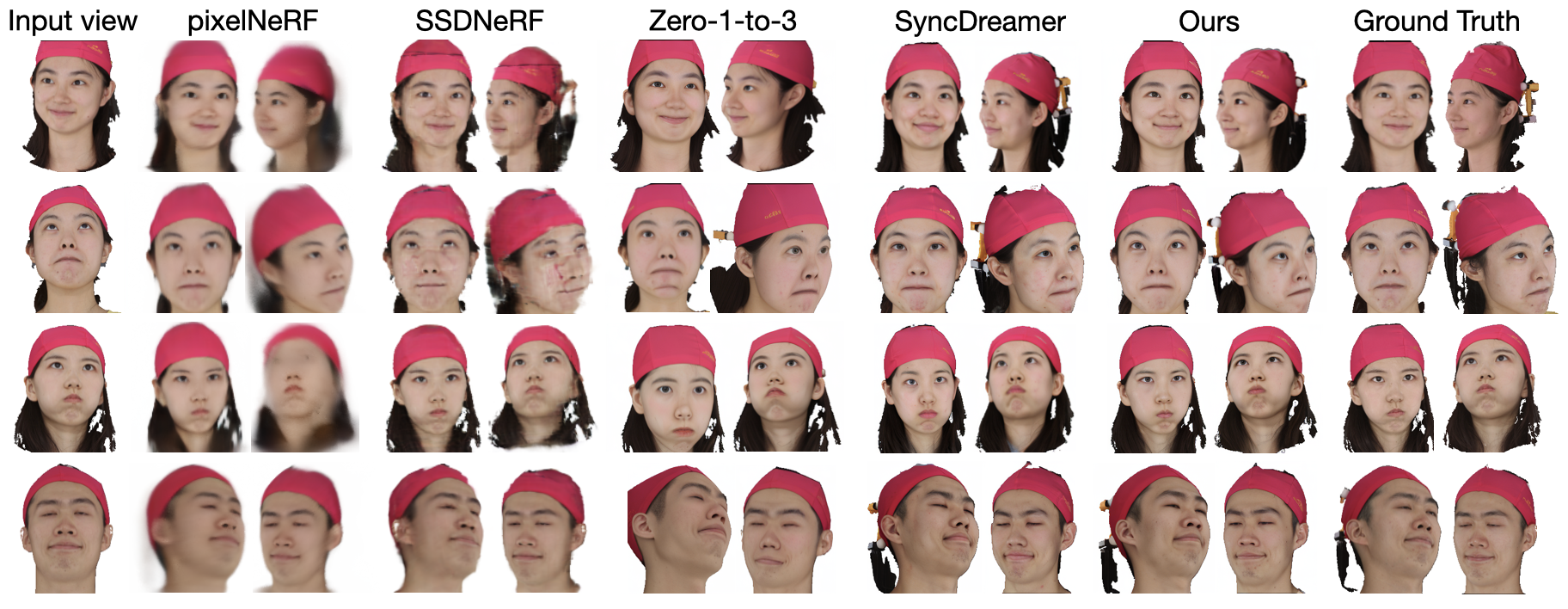}
\caption{\textbf{Single-view reconstruction of human faces.} In addition to the single input view, our method also takes as input a mesh of the facial expression corresponding to the input image. Our method produces more plausible and realistic novel views compared to state-of-the-art methods. While providing multi-view consistency, PixelNeRF \cite{yu2020pixelnerf} and SSDNeRF \cite{ssdnerf} produce overly blurry results. Zero-1-to-3 \cite{Liu_2023_ICCV} generates images of good quality which however fail to preserve multi-view consistency and do not align with the ground truth target views. SyncDreamer~\cite{liu2023syncdreamer} produces multi-view consistent images with relatively accurate facial expressions that however lose the resemblance. 
For more details and discussion, please see section \ref{subsec:nvs}.}
\label{fig:novel_view_synthetis}
\end{figure*}

\section{Experiments}
We demonstrate the effectiveness of our method on the novel view synthesis (Sec.~\ref{subsec:nvs}) and the animation from a single image (Sec.~\ref{subsec:novel_expression}). 
For ablation studies on design choices and training regimes, and discussion about the effect of the topology and expressiveness of the input meshes, please refer to the supplementary material.

\boldparagraph{Training} 
To train our model, we exploit diffusion models Zero-1-to-3~\cite{Liu_2023_ICCV} and SyncDreamer~\cite{liu2023syncdreamer} that have been trained on a large dataset such as Objaverse \cite{objaverse} (800k objects). 
To further allow better generalization to any user-input images, we remove the input view parameter embeddings by setting all the corresponding values to zeros. Additionally, we drop the assumption that the input and the first target image have the same azimuth.

\begin{figure}[h]
\centering
\includegraphics[width=\linewidth]{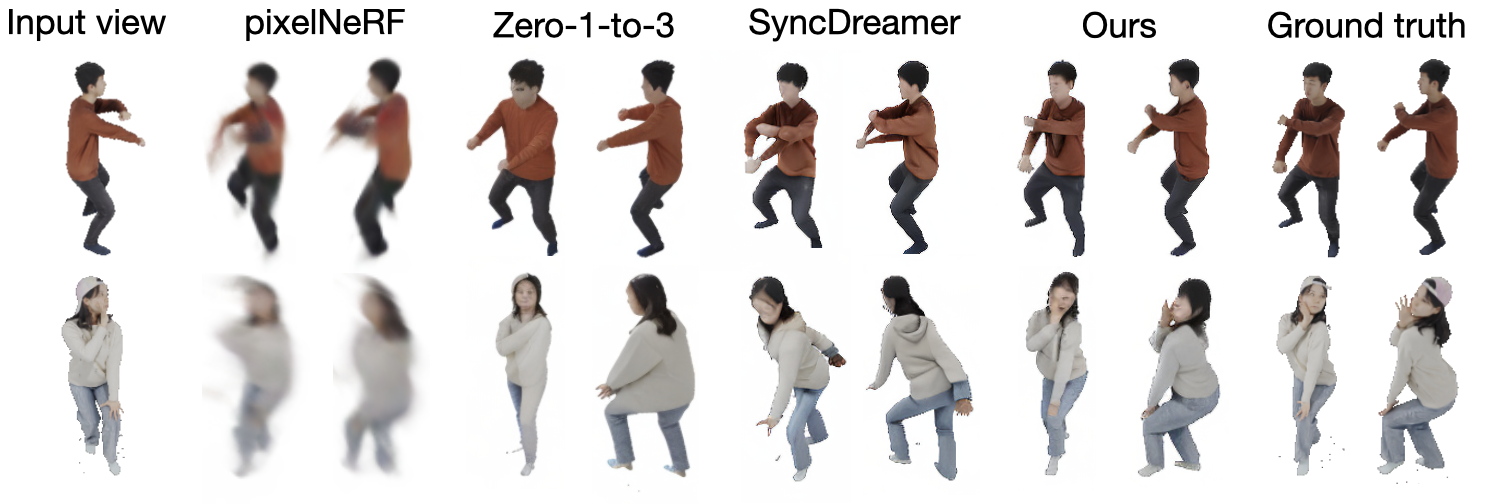}
\caption{\textbf{Single-view reconstruction of human bodies.} Our method is the only one that reconstructs the correct body poses. The relatively low resolution of all methods, however,
limits the amount of details in the generated images.}
\label{fig:full_body}
\end{figure}

\begin{figure*}
\centering
\includegraphics[width=\textwidth]{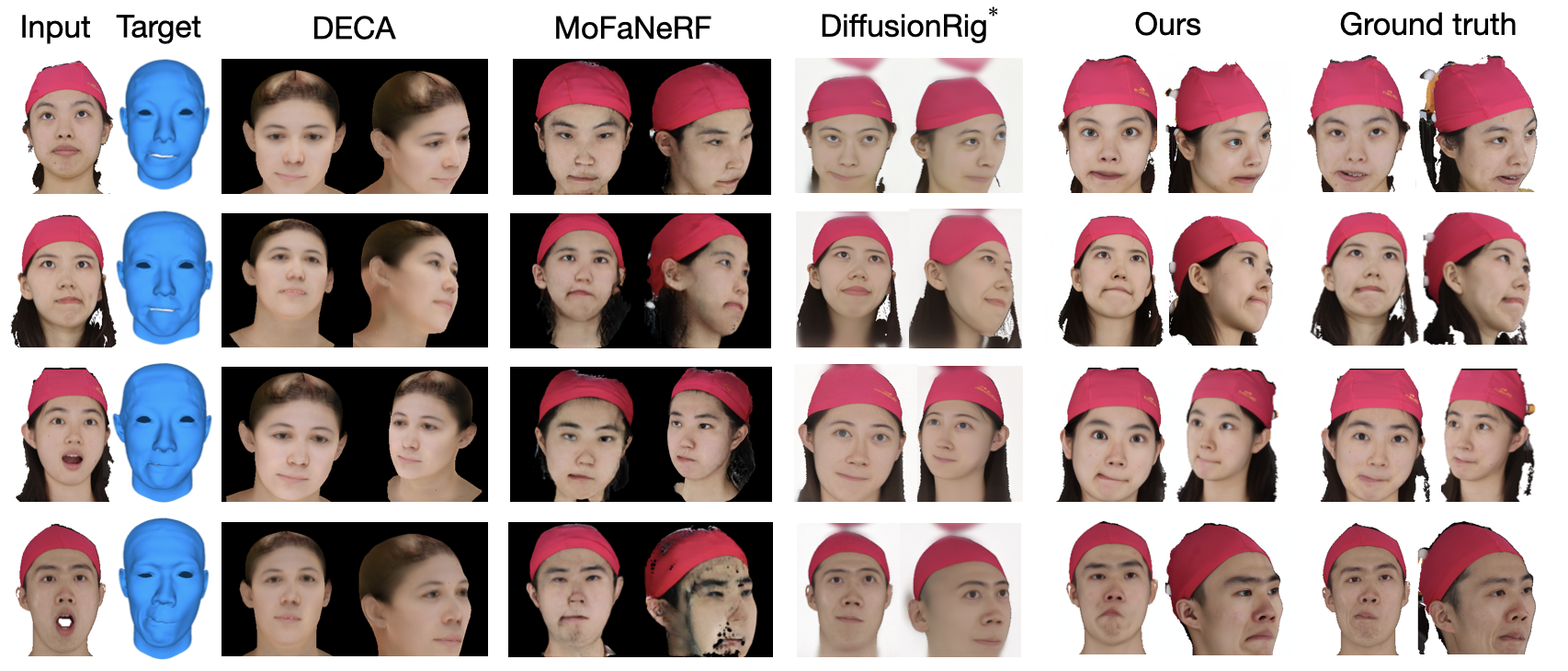}
\caption{\textbf{Novel facial expression synthesis.} Qualitative comparison with DECA\cite{Feng:SIGGRAPH:2021}, MoFaNeRF \cite{zhuang2022mofanerf}, and DiffusionRig \cite{ding2023diffusionrig} on novel facial expression synthesis. DiffusionRig is denoted with $^{*}$ since it requires per-subject finetuning with additional images. Our morphable diffusion model is the only one that successfully synthesizes novel views for a novel facial expression while retaining high fidelity. For more details and discussion, please see section \ref{subsec:novel_expression}.}
\label{fig:novel_expression}
\end{figure*}

\boldparagraph{Experimental setup}
To evaluate our model, we use a dataset of human faces (FaceScape~\cite{yang2020facescape}) and bodies (THuman 2.0 \cite{tao2021function4d}). 
For the novel view synthesis task of the human faces, 
we train our model on 323 subjects, each having 20 unique facial expressions, and evaluate its performance on the remaining 36 subjects, including the 4 subjects who agree to be shown in publications. 
The results of these 4 subjects are displayed in our qualitative evaluation figures. Since the capture devices are visible from the back views, we only include views that have azimuth in the range of [-90, 90] degrees for our training and evaluation. 
During testing, we use input images that have absolute azimuth and elevation both less than 15 degrees, since using more extreme facial poses does not provide a sufficient amount of information for any meaningful reconstruction. 
We follow DINER \cite{prinzler2022diner} and apply cropping, rescaling, and color calibration. 
FaceScape provides coarse meshes of the fitted bilinear model for each frame, which we use to build the voxel grids for SparseConvNet (Sec.~\ref{subsec:MorphableDiffusion}). 

For the novel view synthesis of full bodies, we evaluate our method on the THuman 2.0 dataset \cite{tao2021function4d}, which is a synthetic dataset that consists of 526 meshes with ground truth SMPL-X \cite{SMPL-X:2019} fittings. 
We follow \cite{liu2023syncdreamer} and render 16 input and 16 target views (with the same azimuth) for each mesh in a spherical camera path with an elevation of 30 degrees. 
The SMPL-X vertices for each mesh are used to build voxel grids and query the SparseConvNet. We use the first 473 meshes for training and the last 53 meshes for testing.
\label{sec:experiments}

\boldparagraph{Metrics} We compare SSIM \cite{ssim}, LPIPS \cite{zhang2018perceptual}, Frechet Inception Distance (FID) \cite{fid_nips}, Percentage of correct Keypoints (PCK) \cite{mpii}, and face re-identification accuracy (Re-ID) \cite{phillips2000feret} on the generated images of our method and the baselines. We use an off-the-shelf facial keypoints regressor \cite{mmpose2020} with HRNetV2 backbone \cite{hrnetv2} to predict 68 facial keypoints and normalize them using the intercanthal distance between two eyes. %
Due to the slightly inaccurate facial expressions in the ground truth bilinear meshes, we also run the keypoint regressor on ground truth images and use the predicted keypoints as pseudo ground truth, instead of using the projected 3D keypoints from the bilinear meshes.
We then label the keypoints predicted from the generated face images that are within 0.2 pixels away from the pseudo ground truth as correct ones. Note that we only evaluate PCK for camera poses that have both absolute azimuth less than 30 degrees and absolute elevation less than 15 degrees, in order to avoid potential failure cases of the keypoint regressor on larger angles where some keypoints are invisible. We also follow \cite{ding2023diffusionrig} to report face re-identification accuracy, where we extract 128-dimensional face descriptor vectors from the face images using the dlib library \cite{dlib}. 
If two vectors between ground truth and the generated image have an Euclidean distance of less than 0.6, we classify the generated image to be re-identified as the same person. 
In addition, we reconstruct meshes by training NeuS2 \cite{neus2} on the generated images. We show more results on mesh reconstruction in the supplementary.

\subsection{Novel view synthesis} \label{subsec:nvs}
\begin{table}%
\centering
\scalebox{0.73}{
\begin{tabular}{c|c|c|c|c|c}

Method      & LPIPS ↓  & SSIM ↑ & FID ↓ & PCK@0.2 ↑ & Re-ID ↑  \\ \hline
pixelNeRF \cite{yu2020pixelnerf}   &     0.2200   &   0.7898           &           92.61      &   72.34  & 97.46  \\
Zero-1-to-3 \cite{Liu_2023_ICCV}   & 0.4248       &  0.5656             &   10.97              &  4.69 & 96.77    \\ 
SSD-NeRF \cite{ssdnerf} & 0.2225 & 0.7225 & 34.88 & 92.65 & 98.74 \\
SyncDreamer \cite{liu2023syncdreamer}        &   {0.1854}           &   0.7732             & \textbf{6.05} & 94.07 & 99.60   \\ 
\hdashline
Ours        &  \textbf{0.1653}      &   \textbf{0.8064}    &     6.73             & \textbf{95.80}  & \textbf{99.86}    \\ \hline
\end{tabular}}
\caption{\textbf{Novel view synthesis of human faces.} Quantitative comparison on the FaceScape \cite{yang2020facescape} dataset demonstrates that our method produces more realistic face images with more accurate facial expressions and better resemblance.}
\label{tab:facescape-novel-view}
\end{table}

\begin{table}%
\centering
\scalebox{0.92}{
\begin{tabular}{c|c|c|c}

Method      &LPIPS ↓  & SSIM ↑ & FID ↓  \\ \hline
pixelNeRF \cite{yu2020pixelnerf}   & 0.1432       & 0.8759       & 104.42                          \\ 
Zero-1-to-3 \cite{Liu_2023_ICCV}   & 0.1163       & 0.8764       & 49.94                        \\ 
SyncDreamer \cite{liu2023syncdreamer}        & 0.0960       &  0.8826       & 41.33                \\ 
\hdashline
Ours        & \textbf{0.0625}       & \textbf{0.9181}       &  \textbf{30.25}                           \\ \hline
\end{tabular}}
\caption{\textbf{Novel view synthesis of full human bodies.} Quantitative comparison for full-body novel view synthesis on the THuman 2.0 \cite{tao2021function4d} dataset. Our method demonstrates a considerable improvement over all the baselines across all metrics. }
\label{tab:full-body}
\end{table}

\boldparagraph{Baselines} 
We adopt pixelNeRF \cite{yu2020pixelnerf}, Zero-1-to-3 \cite{Liu_2023_ICCV}, SyncDreamer \cite{liu2023syncdreamer}, and SSD-NeRF \cite{ssdnerf} as baselines for the novel view synthesis task. For fair comparisons, we finetune the baselines or train them from scratch on our training data. Note that although SyncDreamer proposes not to finetune the UNet along with its conditioning module, we find it beneficial to do so when we transfer the domain of the SyncDreamer model pretrained on general objects to human face / bodies. Therefore, the UNet is also finetuned in the baseline SyncDreamer models that we report in this paper. We refer to the supplementary for the advantages of finetuning the UNet and further details on training the baseline methods, as well as an additional qualitative evaluation of novel view synthesis of faces with EG3D \cite{chan2022efficient}.

\boldparagraph{Novel view synthesis of faces} 
Fig.~\ref{fig:novel_view_synthetis} and Tab.~\ref{tab:facescape-novel-view} show the qualitative and quantitative results of our method versus the baselines for novel view synthesis on the FaceScape test subjects. 
PixelNeRF and SSD-NeRF both preserve the resemblance, however, produce blurry results. 
SyncDreamer generates views with good multi-view consistency and reconstructs relatively accurate facial expressions, but sometimes fails to preserve the resemblance. Zero-1-to-3 preserves resemblance to some extent, but the generated results are slightly distorted and are misaligned with the target views, which causes inferior results on the structural and keypoint metrics. This is caused by the fact that the method assumes a uniform camera intrinsic matrix for all data; this way, the varying intrinsic camera information in the considered dataset is never taken into account by the model.
Our method produces the best scores on most of the metrics while preserving good resemblance, which is attributed to the effective conditioning on the morphable model.

\boldparagraph{Novel view synthesis of full bodies} 
Fig. ~\ref{fig:full_body} and Tab.~\ref{tab:full-body} demonstrates more significant improvement of full body reconstructions using our method compared to the facial data, indicating the necessity of a human prior in a diffusion pipeline. 
However, the generated full body images for all methods lack details due to the relatively low resolutions.

\begin{figure*}[t]
\centering
\includegraphics[width=1.0\textwidth]{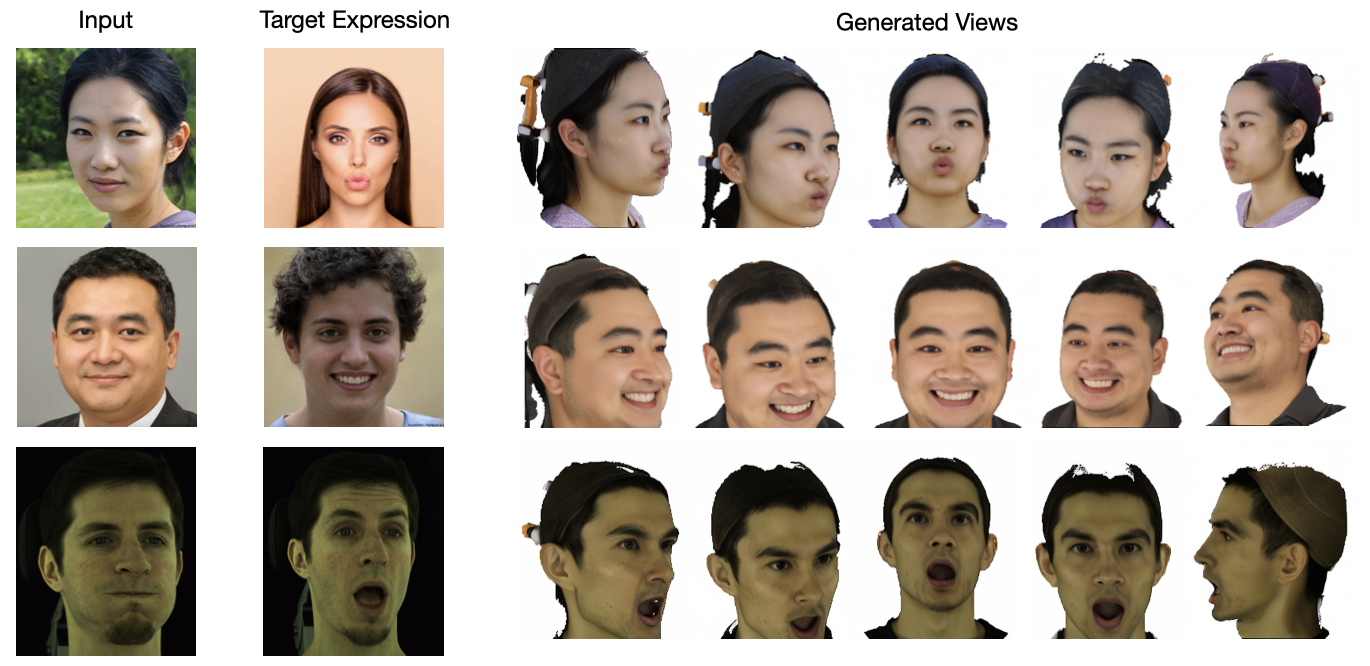}
\caption[caption]{
\textbf{Generalization abilities.} We show the results of our novel facial expression synthesis model on two Asian subjects generated by StyleGAN2 \cite{Karras2019stylegan2} and one subject from the Multiface dataset \cite{wuu2022multiface}. Our model is capable of maintaining 3D consistency of the generated imagery and preserving target facial expression. However, it struggles with the hairstyle reconstruction and out-of-distribution ethnicities due to the strong bias of the training data towards hair caps and Asian subjects. We use MICA \cite{MICA:ECCV2022} to optimize for the shape and expression parameters of the FLAME model from the input and target expression images, respectively. All results in this figure are generated using the model trained with FLAME fittings. For more details about models trained with bilinear / FLAME topologies, please refer to the supplementary. The target expression image in the first row is courtesy of © Deagreez / iStock.
}
\label{fig:generalization}
\end{figure*}

\subsection{Novel facial expression synthesis} \label{subsec:novel_expression} 
We further evaluate our method for multi-view consistent novel facial expression synthesis.
We hold out the ``jaw\_right" expression in the FaceScape dataset from training and train our model on the remaining 19 expressions.

\boldparagraph{Baselines} Using the same metrics as for novel view synthesis, we compare our method with 
DiffusionRig \cite{ding2023diffusionrig}, DECA \cite{Feng:SIGGRAPH:2021}, and MoFaNeRF \cite{zhuang2022mofanerf}.
Note that DiffusionRig requires per-subject finetuning with additional images and is thus denoted with the $^{*}$ symbol. For this baseline, we randomly choose 20 views from the 20 expressions and finetune the second stage for 5k steps for each subject.
For a fair comparison to DECA, we fit FLAME \cite{FLAME:SiggraphAsia2017} parameters to the ground truth meshes of the test expression in FaceScape's bilinear topology, instead of using the predicted ones. We then simply render the predicted albedo maps onto the ground truth meshes and use them to render albedo and lambertian images for 2D conditioning in DiffusionRig, 
Although the pre-trained model of MoFaNeRF has already been trained on all 20 facial expressions and some of our test subjects in the FaceScape dataset, we still treat the input image in a random facial expression as a novel subject and finetune the pre-trained model for 2k steps (as suggested by the authors).

\boldparagraph{Disentanglement of reconstruction and animation} 
We propose an efficient training scheme that disentangles the reconstruction (guided by the input image) and the animation (guided by the underlying morphable model). At every training step, given an input image of a subject, we randomly sample a novel facial expression from the dataset that does not align with the expression of the input image. 
This cross-expression training, besides allowing explicit control of generated images, further improves the training efficiency as it serves as a data augmentation procedure and reduces potential overfitting on small datasets. 
At inference time, we could either use the expression that corresponds to the input image to enable novel view synthesis or use a novel expression to animate the reconstruction. We refer the readers to the ablation studies section in the supplementary for more discussion about the effectiveness of this proposed training regime.

\begin{table}%
\centering
\scalebox{0.72}{
\begin{tabular}{l|c|c|c|c|c}

 &LPIPS ↓  & SSIM ↑ & FID ↓ & PCK@0.2 ↑ & Re-ID ↑ \\ \hline

DiffusionRig* \cite{ding2023diffusionrig}   &      0.2534  &  0.7438      & 42.93        &    89.74  & 97.67   \\
DECA \cite{Feng:SIGGRAPH:2021}   & 0.3393 & 0.6904    &  182.25        &   91.06  & 25.23                \\ 
MoFaNeRF \cite{zhuang2022mofanerf}  & 0.2877 &   0.6956  &  32.92        & 79.81 &  88.97               \\ \hdashline
Ours    &  \textbf{0.1693}   &  \textbf{0.8026}      &     \textbf{14.34}          &    \textbf{95.46}   & \textbf{99.89} \\
\hline
\end{tabular}}
\caption{\textbf{Novel expression synthesis.} Quantitative evaluation for novel expression synthesis on the held-out ``jaw\_right" expression. 
When evaluating against ground truth images with a white background, we convert the background pixels to white for all baselines using the ground truth masks. Our method outperforms the baselines on all metrics.}
\label{tab:facescape_novel_expression}
\end{table}

Fig.~\ref{fig:novel_expression} and Tab.~\ref{tab:facescape_novel_expression} show the qualitative and quantitative results of our method versus the baselines on the novel expression synthesis task. DiffusionRig is unable to control eye poses since DECA that it depends on for mesh reconstruction during the first stage training does not model eye poses. It also generates slightly distorted faces in extreme head poses. DECA renders meshes with the predicted albedo maps, which is unable to produce photorealistic facial details.
Expression rigging with MoFaNeRF loses resemblance and produces artifacts when the input facial expressions are not neutral. Our method, on the other hand, is able to generate realistic facial images under novel expressions from various views while retaining high visual quality, with input image in any facial expression.

\section{Limitations and future work} \label{subsec:limitations}

While our morphable model exhibits promising capabilities, it is essential to acknowledge its inherent limitations that could impact its widespread applicability. As shown in Fig.~\ref{fig:generalization}, one significant constraint arises from the current inability to generalize effectively across various ethnicities and hair types, primarily stemming from the constraints of the FaceScape dataset, which predominantly features Asian subjects wearing a distinctive red cap. This limited diversity poses a challenge in training the model to handle the rich variability present in real-world scenarios, where individuals showcase diverse ethnicities and hair textures, lengths, and styles. We believe that this limitation can be alleviated with the advent of more diverse head datasets, such as RenderMe-360 \cite{2023renderme360}.

Additionally, we find that our pipeline does not generalize well to out-of-distribution camera parameters that are significantly different from the ones used during training. Therefore, our current implementation relies on an external NeRF-based method, such as \cite{wang2021neus, neus2}, for comprehensive 3D facial reconstruction and free-path novel view synthesis. While this approach has been proven effective, it introduces a dependency on an external system, potentially impacting the model's standalone usability and flexibility. Improving the generalizability to any camera views and exploring ways to integrate a more self-contained 3D reconstruction process within our model could be a valuable avenue for future research.

Finally, the resolution of our model is limited at 256$\times$256 as we inherit this limitation from the components proposed in the previous works \cite{Liu_2023_ICCV, liu2023syncdreamer}. Rendering full bodies under such a low resolution provides a limited amount of details. Future works could involve training a full-body specific model from scratch at a higher resolution or integrating a super-resolution module into the pipeline.

\section{Conclusion} \label{sec:conclusion}
Our work presents a method for avatar creation through the introduction of a novel morphable diffusion model. By seamlessly integrating a 3D morphable model into a multi-view consistent diffusion framework, we have successfully incorporated a powerful human prior. This serves a dual purpose: firstly, it enhances the finetuning process for generative diffusion models, allowing for effective adaptation from extensive datasets to more limited, human-specific datasets. Secondly, it empowers the direct manipulation of synthesized novel views, offering a level of control and realism that surpasses current methodologies.
Our qualitative and quantitative evaluation demonstrates the superior quality achieved by our model in comparison to state-of-the-art methods. 
We hope that our method will positively contribute to accelerating the field of photorealistic digitization of humans and promote follow-up research to address the current limitations. 

\section*{Acknowledgements} \label{sec:acknowledgements}
We thank Korrawe Karunratanakul for the fruitful discussions, Timo Bolkart for the advice on fitting the FLAME model, and Malte Prinzler for the help with the color-calibrated FaceScape data. This project is partially supported by the SNSF grant 200021 204840. 
SW acknowledges the ERC Starting Grant LEGO-3D (850533) and the DFG EXC number 2064/1 - project number 390727645.

{
   \small
   \bibliographystyle{ieeenat_fullname}
   \bibliography{main}
}

\clearpage
\appendix
\numberwithin{equation}{section}
\setcounter{equation}{0}
\numberwithin{figure}{section}
\setcounter{figure}{0}
\numberwithin{table}{section}
\setcounter{table}{0}
\maketitlesupplementary

In this supplementary document, we first provide additional evaluations of the novel view and expression synthesis of faces (Sec.~\ref{sec:face_nvs_supp} \&~\ref{sec:face_nes_supp}). 
We further provide details for our implementation (Sec.~\ref{sec:impl_details_supp}) and comparison to baseline methods (Sec.~\ref{sec:baseline_comparison_supp}). We also ablate our model for different design choices and training scheme (Sec.~\ref{sec:ablation}). Finally, we discuss the effect of the topology, expressiveness, and accuracy of the input meshes on the generated face images (Sec.~\ref{sec:mesh}).

\section{Novel view synthesis of faces}
\label{sec:face_nvs_supp}

\begin{figure*}[h]
\centering
\includegraphics[width=\textwidth]{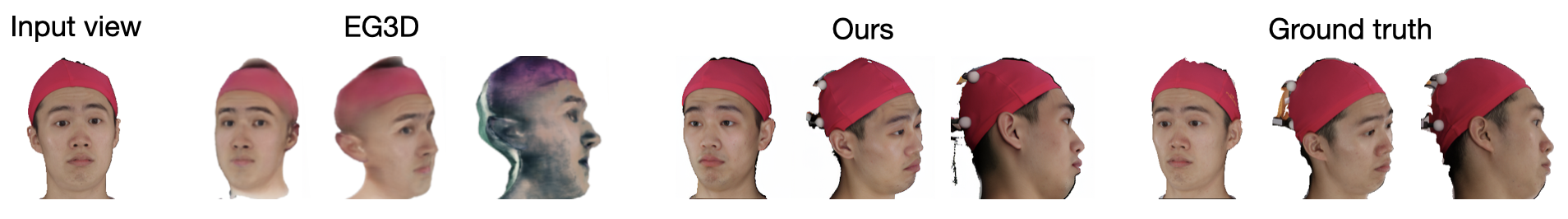}
\caption{\textbf{Additional qualitative results on novel view synthesis of faces.} 
EG3D \cite{chan2022efficient} fails to generate side views with good quality, while our method generates high-fidelity images in all views.}
\label{fig:face_nvs}
\end{figure*}

\begin{figure*}[h]
\centering
\includegraphics[width=\textwidth]{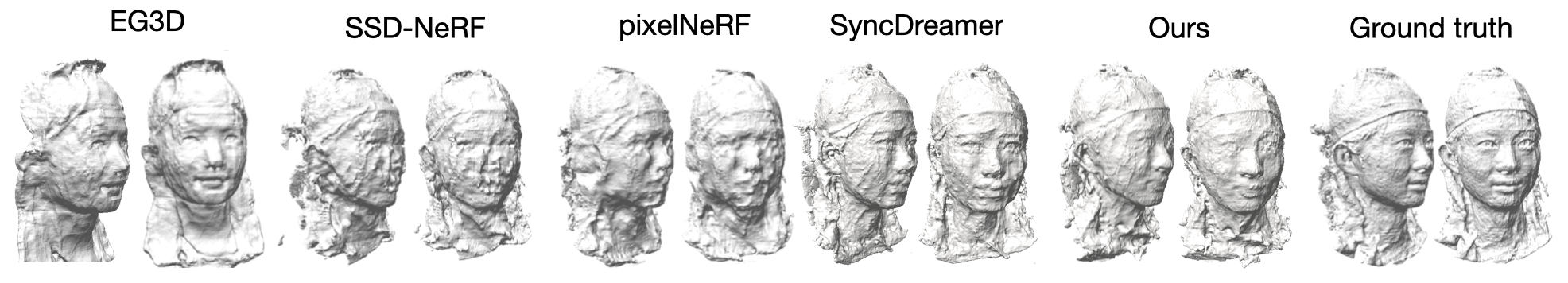}
\caption{\textbf{Mesh reconstruction of faces.} EG3D \cite{chan2022efficient} generates bad geometries for side views. SSD-NeRF \cite{ssdnerf} and pixelNeRF produce overly coarse geometries. SyncDreamer \cite{liu2023syncdreamer} and our method produces meshes with a comparable amount of details, but our method preserves better facial resemblance (Fig. 3 in the paper) and has the additional capability for expression rigging.}
\label{fig:face_mesh_nvs}
\end{figure*}

\boldparagraph{Comparison to EG3D} We conduct an additional evaluation to compare our method with EG3D \cite{chan2022efficient} for novel view synthesis on the FaceScape dataset \cite{yang2020facescape}. Fig.~\ref{fig:face_nvs} shows the qualitative results. 
The novel view synthesis with the optimized latent code obtained by GAN inversion of EG3D from the input view does not preserve the identity well and fails to generate realistic side views. This can be largely attributed to the fact that the model is trained on mostly frontal views. 
Note that since EG3D requires fixed camera distance, and the scale of the face on the pre-trained model is not available, it is not possible to use the real camera parameters for novel view synthesis. Instead, we render images with their camera parameters and use NeuS2 \cite{neus2} for mesh reconstruction and novel view synthesis. However, it is still challenging to align the EG3D meshes accurately with the ground truth. Therefore, we do not perform a quantitative evaluation for this method and only show the qualitative results in the figure.

\boldparagraph{Geometry evaluation} In addition, we reconstruct the meshes for all the baselines and our method using NeuS2 \cite{neus2} and compare the geometry quality. Since the hairstyle sometimes varies among each generated batch of 16 views in our method, we only reconstruct the mesh using one of the batches of generated images. For a fair comparison, we only use the first 16 generated views to reconstruct the meshes for all baselines. Instead of using the provided ground truth mesh scans in the FaceScape dataset, we also reconstruct the ground truth meshes with NeuS2, using all ground truth target views whose absolute camera azimuth is less than 90 degrees. We consider this evaluation strategy since some parts of the capture devices are visible in the ground truth mesh scans, but not in the reconstructed meshes of any of the compared methods.

\begin{table}[h]
\centering
\scalebox{0.95}{
\begin{tabular}{l|c|c}
Method  & Chamfer Distance ↓ & Volume IoU ↑ \\ \hline

zero-1-to-3 \cite{Liu_2023_ICCV}        & 0.0950            & 0.0613           \\
SyncDreamer \cite{liu2023syncdreamer} &  0.0138            &  0.7947          \\
SSD-NeRF \cite{ssdnerf} &    0.0154          &  0.7801          \\
pixelNeRF \cite{yu2020pixelnerf}     &   \colorbox{Salmon}{0.0118}           &  \colorbox{Salmon}{0.8218}          \\ \hdashline
Ours         & \colorbox{Apricot}{0.0130}             &   \colorbox{Apricot}{0.8048}         \\ \hline
\end{tabular}}
\caption{\textbf{Quantitative evaluation of geometry for novel view synthesis of faces.} Colors denote the \colorbox{Salmon}{1st} and \colorbox{Apricot}{2nd} best-performing models. See Sec.~\ref{sec:face_nvs_supp} for details.}
\label{tab:mesh-nvs}
\end{table}

Fig.~\ref{fig:face_mesh_nvs} and Tab.~\ref{tab:mesh-nvs} display the qualitative and quantitative results of the mesh reconstructions of faces. We report the Chamfer Distance and Volume IoU \cite{mescheder2019occupancy} for geometry accuracy. Even though our meshes demonstrate good geometric details, the quantitative scores are slightly outperformed by pixelNeRF, which however produces overly coarse meshes due to its blurry rendering results. We suspect the main reason to be that our model sometimes generates a hairstyle that is different from the one in the input image, \eg, long hair for male subjects. Overall, the meshes reconstructed from our method are visually comparable to the ones from SyncDreamer. However, our method preserves better resemblance, as shown in Fig. 3 of the paper, and has the additional advantage of the capability of facial expression rigging over SyncDreamer.

\section{Novel facial expression synthesis}
\label{sec:face_nes_supp}

\begin{figure*}[h]
\centering
\includegraphics[width=\linewidth]{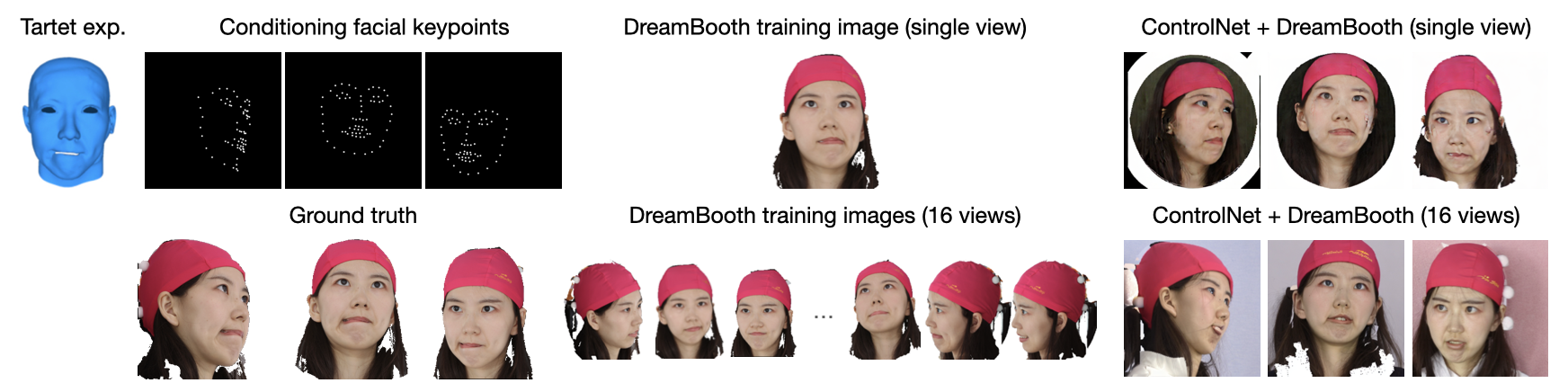}
\caption{\textbf{Facial expression rigging with ControlNet \cite{zhang2023adding} on personalized DreamBooth \cite{ruiz2022dreambooth} models.} The model trained with a single view overfits to the expression in the training image, while the model trained with multiple views in different facial expressions fails to generate the correct facial expression based on the provided conditioning facial keypoints maps.}
\label{fig:controlnet}
\end{figure*}

\boldparagraph{Expression rigging with ControlNet} We test the ability of ControlNet \cite{zhang2023adding} of facial expression rigging by using the OpenPose \cite{openpose} model for human pose conditioning and the projected ground truth facial keypoints as input image. We first personalize the Stable Diffusion 1.5 model \cite{rombach2022high} with DreamBooth \cite{ruiz2022dreambooth} using one or multiple images of the test subject. Fig. \ref{fig:controlnet} shows that the personalized model finetuned on the single input view tends to overfit to the specific facial expression in the input image. The model finetuned with 16 views of the test subject in random views and random facial expressions, however, fails to generate images with the correct facial expressions conditioning merely on the facial keypoint maps. We believe the main reason to be that the current human pose conditioning model isn't trained with various facial expressions. A specific facial expression conditioning model with facial keypoint maps as inputs could potentially enhance the ability of expression rigging.

\begin{table}[h]
\centering
\scalebox{0.9}{
\begin{tabular}{l|c|c}
\textit{Single image input:}   & Chamfer Distance ↓ & Volume IoU ↑ \\ \hline
MoFaNeRF \cite{zhuang2022mofanerf}     & 0.0234             &  0.7281          \\
DECA \cite{Feng:SIGGRAPH:2021}        & 0.0279             &  0.6818          \\ 
DiffusionRig* \cite{ding2023diffusionrig} &  0.0284            &   0.6383         \\ \hdashline
Ours & \textbf{0.0113} & \textbf{0.7670} \\ \hline

\end{tabular}}
\caption{\textbf{Quantitative evaluation of geometry for novel facial expression synthesis.} }
\label{tab:mesh-nes}
\end{table}

\begin{figure*}[h]
\centering
\includegraphics[width=\textwidth]{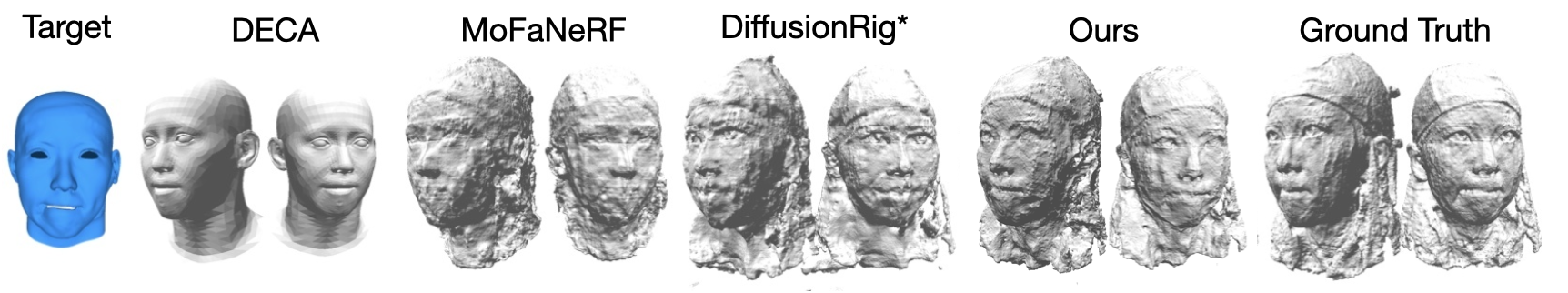}
\caption{\textbf{Mesh Reconstruction for novel expression synthesis.} DiffusionRig requires per-subject finetuning with additional images and is thus denoted with the $^{*}$ symbol. For DECA \cite{Feng:SIGGRAPH:2021}, we simply show the pseudo ground truth FLAME \cite{FLAME:SiggraphAsia2017} mesh since it only renders the predicted albedo map onto this coarse mesh. MoFaNeRF \cite{zhuang2022mofanerf} and DiffusionRig \cite{ding2023diffusionrig} produce overly coarse geometries. Our method generates the highest amount of details and preserves the best geometry of the target subject.}
\label{fig:face_mesh_exp}
\end{figure*}

\boldparagraph{Geometry evaluation} Fig.~\ref{fig:face_mesh_exp} and Tab.~\ref{tab:mesh-nes} show the qualitative and quantitative results of the mesh reconstructions on face novel expression synthesis. 
In this case, our method outperforms the baselines in terms of both visual qualities and geometry metrics.

Overall, we believe that although the precise geometry evaluation metrics are well suited for examining the reconstruction fidelity of multi-view  NeRF-based methods, they are less reflective of the model performance in the case of generative models working in highly underconstrained single-image setups.

For the animation of the meshes and textures for all methods, please refer to our project website.

\begin{figure}[h]
\centering
\includegraphics[width=\linewidth]{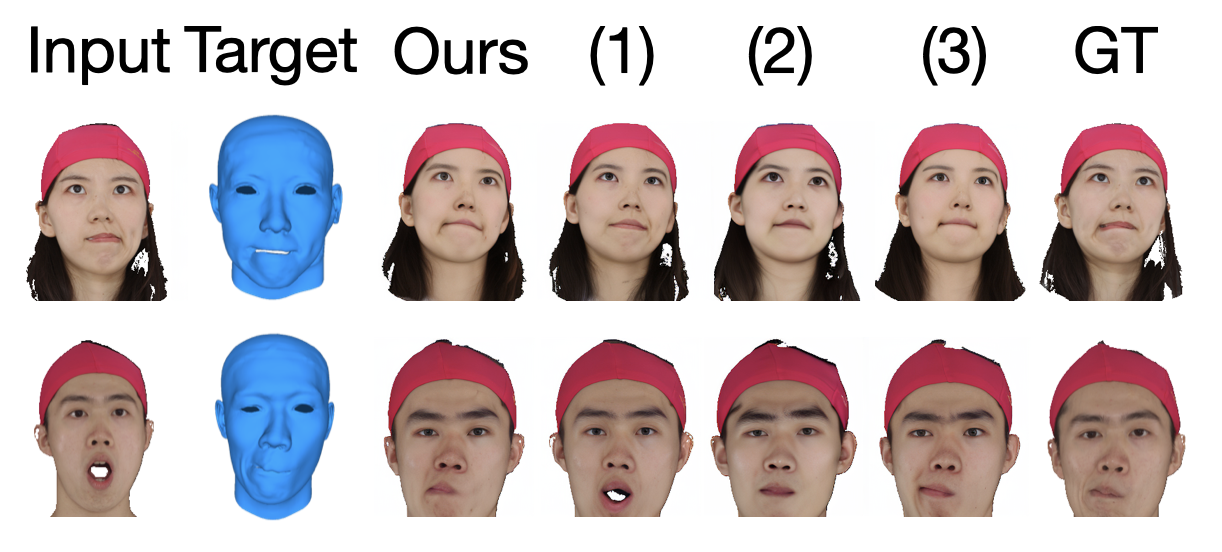}
\caption{\textbf{Ablation studies of our proposed model v.s. different design choices and train-
ing strategies}. Ablated models correspond to training with same facial expression for input and target views (1), not finetuning the UNet along with the conditioning module (2), and training with FLAME meshes fitted to 3D keypoints instead of the ground truth bilinear meshes (3). Model (1) overfits to the input views and tends to ignore the target facial expressions. Model (2) generates faces with worse resemblance. Model (3) generates faces in a comparable resemblance but slightly worse facial expressions due to the optimization loss of the FLAME model.}
\label{fig:ablation}
\end{figure}

\section{Implementation details}
\label{sec:impl_details_supp}

\begin{table}[h]
\scalebox{0.91}{
\begin{tabular}{l|c|c}
\multicolumn{1}{c|}{Layers} & Layer Description            & Output Dim \\ \hline
1-2                             & 2 $\times$ (3 × 3 × 3 conv, stride=1) &  D$\times$H$\times$W$\times$16                \\
3                               & 3 × 3 × 3 conv, stride=2     &  \sfrac{1}{2}D$\times$\sfrac{1}{2}H$\times$\sfrac{1}{2}W$\times$32                \\
4-5                             & 2 $\times$ (3 × 3 × 3 conv, stride=1) & \sfrac{1}{2}D$\times$\sfrac{1}{2}H$\times$\sfrac{1}{2}W$\times$32    \\
6                               & 3 × 3 × 3 conv, stride=2     &  \sfrac{1}{4}D$\times$\sfrac{1}{4}H$\times$\sfrac{1}{4}W$\times$64   \\
7-9                             & 3 $\times$ (3 × 3 × 3 conv, stride=2) &  \sfrac{1}{4}D$\times$\sfrac{1}{4}H$\times$\sfrac{1}{4}W$\times$64   \\ \hline
\end{tabular}}
\caption{\textbf{Architecture of our SparseConvNet.}}
\label{tab:architecture}
\end{table}

We train our model with the AdamW~\cite{loshchilov2017decoupled} optimizer and a total batch size of 140 images for 6k steps ($\approx$36 hours) using two 80GB NVIDIA A100 GPUs. The learning rate for training the UNet is increased from 1e-6 to 5e-5 with 100 warm-up steps \cite{goyal2017accurate}. The learning rate for the remaining trainable modules are set to 5e-4. During each training step, we randomly select 1 view as input and $N=16$ target views. 
For inference, our method takes about 25 seconds to generate 16 target views from a single input image using 50 DDIM \cite{song2020denoising} steps with an NVIDIA RTX 3090 GPU. 

Tab.~\ref{tab:architecture} shows the architecture of our SparseConvNet $f_\theta$ (introduced in Section 3.2 in the paper). Given vertex features $\textbf{V}_F \in \mathbb{R}^{n_v \times d}$, where $n_v$ is the number of vertices and $d$ is the dimensionality of the noise features, a sparse volume $\textbf{V}_S \in \mathbb{R}^{D \times H \times W \times d}$ filled with the sparse vertex features is first constructed. Here, $D, H, W$ are determined by the size of the bounding box of the face/full-body mesh, which differs for each mesh. Then, the SparseConvNet downsamples the size of the volume by 4 times, while increasing the number of channels to $f_V$ by 4 times, and trilinearly interpolates the 3DMM-aware feature grid $\mathbf{F}_V$ from the downsampled volume. 
The SparseConvNet is implemented using the Spconv library \cite{spconv2022}.

In practice, we upsample the noisy image features to 16 channels with a 2D CNN block pretrained in SyncDreamer, and unproject and interpolate these noise features to construct $\mathbf{V}_F$. Therefore, we have $d=16$ and $f_V=64$. The size of the grid $\mathbf{F}_V$ is set as $x=y=z=32$, and the size of the frustums $\mathbf{F}^{(i)}$ are set as $h_F = w_F = 32$, $d_F = 48$.

We set the voxel size for our SparseConvNet to 0.005
and the length of frustum volume to $\sfrac{\sqrt{3}}{2}$, same as their defaults in SyncDreamer. 

\section{Additional details on baseline comparisons}
\label{sec:baseline_comparison_supp}
For a fair comparison, we finetune zero-1-to-3 and SyncDreamer models pre-trained on Objaverse \cite{objaverse} on the FaceScape / THuman 2.0 datasets, for 6k steps. We finetune the UNet together with the conditioning module for SyncDreamer (see Sec.~\ref{sec:ablation} for the detailed explanation). We train pixelNeRF and SSD-NeRF from scratch on the same datasets for 400k and 80k steps respectively, following the training schemes of the original pipelines. At inference, we use the same single input view for all methods. Finetuning and pre-training are performed with the default hyperparameters of the baseline methods. 
For EG3D, we use a custom GAN inversion repository \cite{eg3dprojector} to optimize the latent codes for 1000 steps for all input images. 

\section{Ablation studies}
\label{sec:ablation}
\begin{table}%
\centering
\scalebox{0.52}{
\begin{tabular}{l|c|c|c|c|c|c}

      &LPIPS ↓  & SSIM ↑ & FID ↓ &  PCK@0.2 ↑ & PCK@0.2 (mouth) ↑ & Re-ID ↑ \\ \hline
Morphable Diffusion    & \textbf{0.1693}       &  \textbf{0.8026}  &   14.34   & \textbf{95.46} & \textbf{94.23} & 99.89                  \\ 
\hdashline
\phantom{+}{\it w.} same expression        & 0.1787       &  0.7881      &  \textbf{13.68}                       & 92.39 & 84.12 & \textbf{100.00}    \\
\phantom{+}{\it w.o.}  finetuning UNet        & 0.1841 & 0.7910 & 20.70 & 93.44 & 90.30 & 98.35  \\
\phantom{+}{\it w.}  FLAME \cite{FLAME:SiggraphAsia2017} meshes        & 0.1764 & 0.7939 & 15.03 & 95.22 & 93.23 & 99.45  \\
\hline
\end{tabular}}

\caption{\textbf{Ablation studies on different design choices and training strategies of our novel facial expression synthesis model.} The proposed pipeline demonstrates superior performance compared to the alternative designs on most metrics. Our proposed model produces the most accurate keypoints for the test facial expression, with the difference even larger for mouth keypoints only.}
\label{tab: ablation}
\end{table}

\begin{table}%
\centering
\scalebox{0.68}{
\begin{tabular}{l|c|c|c|c|c}

      &LPIPS ↓  & SSIM ↑ & FID ↓ &  PCK@0.2 ↑ & Re-ID ↑ \\ \hline
SyncDreamer w. UNet    & \textbf{0.1854}       &  \textbf{0.7732}  &   \textbf{6.05}  & \textbf{94.07} & \textbf{99.60}                  \\ 
SyncDreamer w.o UNet    & 0.2026       &  0.7585  &   7.53   & 88.64 & 96.60   \\
\hline
\end{tabular}}

\caption{\textbf{Ablation studies on the effect of finetuning UNet with SyncDreamer \cite{liu2023syncdreamer} for novel view synthesis on FaceScape \cite{yang2020facescape}.} The model that finetunes the UNet outperforms the model that does not on every metrics, suggesting the need to finetune the UNet when we apply the method to human faces.}
\label{tab:syncdreamer_unet}
\end{table}

We compare our method with several variants of pipeline designs and training strategies: 1) having the same facial expression for input and target views during training, 2) not finetuning the UNet, and 3) instead of using ground truth bilinear meshes provided by the FaceScape dataset \cite{yang2020facescape}, we use FLAME \cite{FLAME:SiggraphAsia2017} meshes fitted to the ground truth 3D keypoints via optimization.
Since the held-out facial expression ``jaw\_right" mainly involves rigging on mouth keypoints, we additionally report the PCK metric on the 20 mouth keypoints, which we denote as ``PCK@0.2 (mouth)".
Tab.~\ref{tab: ablation} and Fig.~\ref{fig:ablation} show that our proposed pipeline quantitatively achieves the best performance on most metrics, while preserving the most accurate facial expressions and resemblance qualitatively.

The ablated model which employs the same facial expressions for both input and target views, exhibits overfitting to the task of novel view synthesis and tends to overlook the driving signal from the target expression mesh. This leads to diminished performance in terms of both image quality and expression preservation metrics. Such observations underscore the effectiveness of our proposed shuffled training scheme. This scheme, by using images with varying facial expressions for the input and target views, successfully disentangles the processes of reconstruction and animation, as elaborated in Section 4.2 of the paper.

Although SyncDreamer \cite{liu2023syncdreamer} proposes not to finetune the UNet when training their proposed conditioning module with a UNet pretrained on the same object dataset, we find that it's still beneficial to do so when we apply this baseline onto the human domain, as shown in Tab.~\ref{tab:syncdreamer_unet}. Therefore, for the baseline SyncDreamer models that we report in the main paper, we finetune the UNet together with the conditioning module. Similarly, we find improvement to finetune the UNet with our proposed method. As shown both qualitatively and quantitatively, our proposed model generates faces with better resemblance compared to the ablated model without finetuning the UNet.

We also fit FLAME model to the ground truth 3D facial keypoints for all meshes in the FaceScape dataset via optimization, and use the FLAME meshes instead of the ground truth bilinear meshes for both training and inference. This model produces faces that are comparable but slightly worse (due to the loss in the mesh-fitting optimization) to the results generated with our proposed model trained and tested with bilinear meshes. We experiment with this model mainly because we find that not changing the mesh topology during inference produces better results, and there are more off-the-shelf FLAME-based methods to reconstruct meshes from single images \cite{Feng:SIGGRAPH:2021,MICA:ECCV2022}. Therefore, we provide this model for better generalization to in-the-wild face images, as shown in Fig. 6 of the paper. More details about mesh topologies will be discussed in Sec.~\ref{sec:mesh}.

\section{Effects of mesh topology, expressiveness, and accuracy}
\label{sec:mesh}

\begin{figure}[h]
\centering
\includegraphics[width=\linewidth]{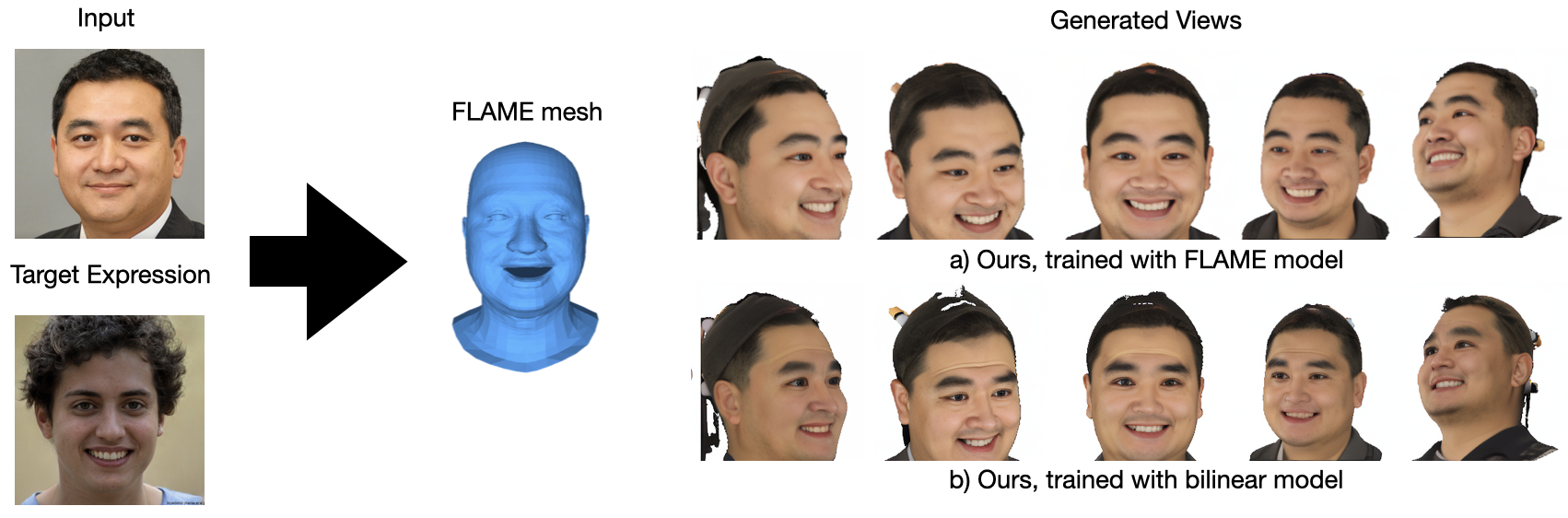}
\caption{\textbf{The effect of changing mesh topology during inference on the generated faces.} a) we test our model trained with FLAME meshes (with 5,023 vertices) on a target expression mesh also in FLAME topology. The shape and facial expression of the mesh are estimated from the input and the target expression images respectively using MICA \cite{MICA:ECCV2022}. b) we use the same input image and mesh but the model is trained with the bilinear mesh topology (with 26,317 vertices). We observe that using the same mesh topology for training / inference leads to results with more accurate facial structures and better resemblance.}
\label{fig:topology}
\end{figure}

\begin{figure}[h]
\centering
\includegraphics[width=\linewidth]{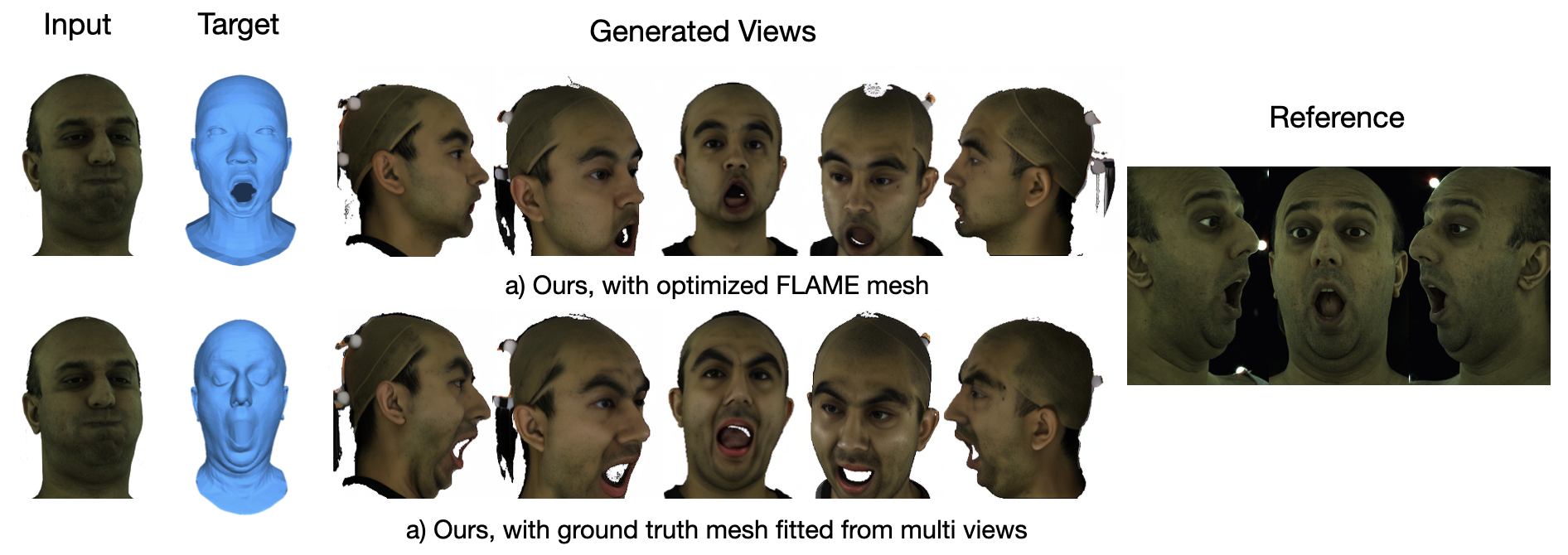}
\caption{\textbf{The effect of the expressiveness of the mesh on the generated faces.} We compare the generated images using a) the FLAME mesh optimized from a single image of the reference expression using MICA \cite{MICA:ECCV2022} v.s. b) using the more expressive ground truth mesh provided in the dataset fitted from all views of the reference expression on a Multiface \cite{wuu2022multiface} subject. We find that using more expressive meshes of the same facial expression improves the resemblance of the generate faces, although neither results preserves good resemblance due to the limited generalization ability of our method on different ethnicities (as discuessed in Sec. 5 of the paper).}
\label{fig:expressiveness}
\end{figure}

Although our method is agnostic of the input mesh topology, thanks to the SparseConvNet's ability to process an arbitrary point cloud, we find that having the same mesh topology for training and inference leads to better results, as shown in Fig.~\ref{fig:topology}. 

However, as shown in Fig.~\ref{fig:expressiveness}, having more expressive meshes could still lead to better preservation of the subject identity. Future works could consider leveraging geometry information of the input image into the conditioning module, such as signed distance fields (SDFs), to improve the resemblance.

\begin{figure}[h]
\centering
\includegraphics[width=\linewidth]{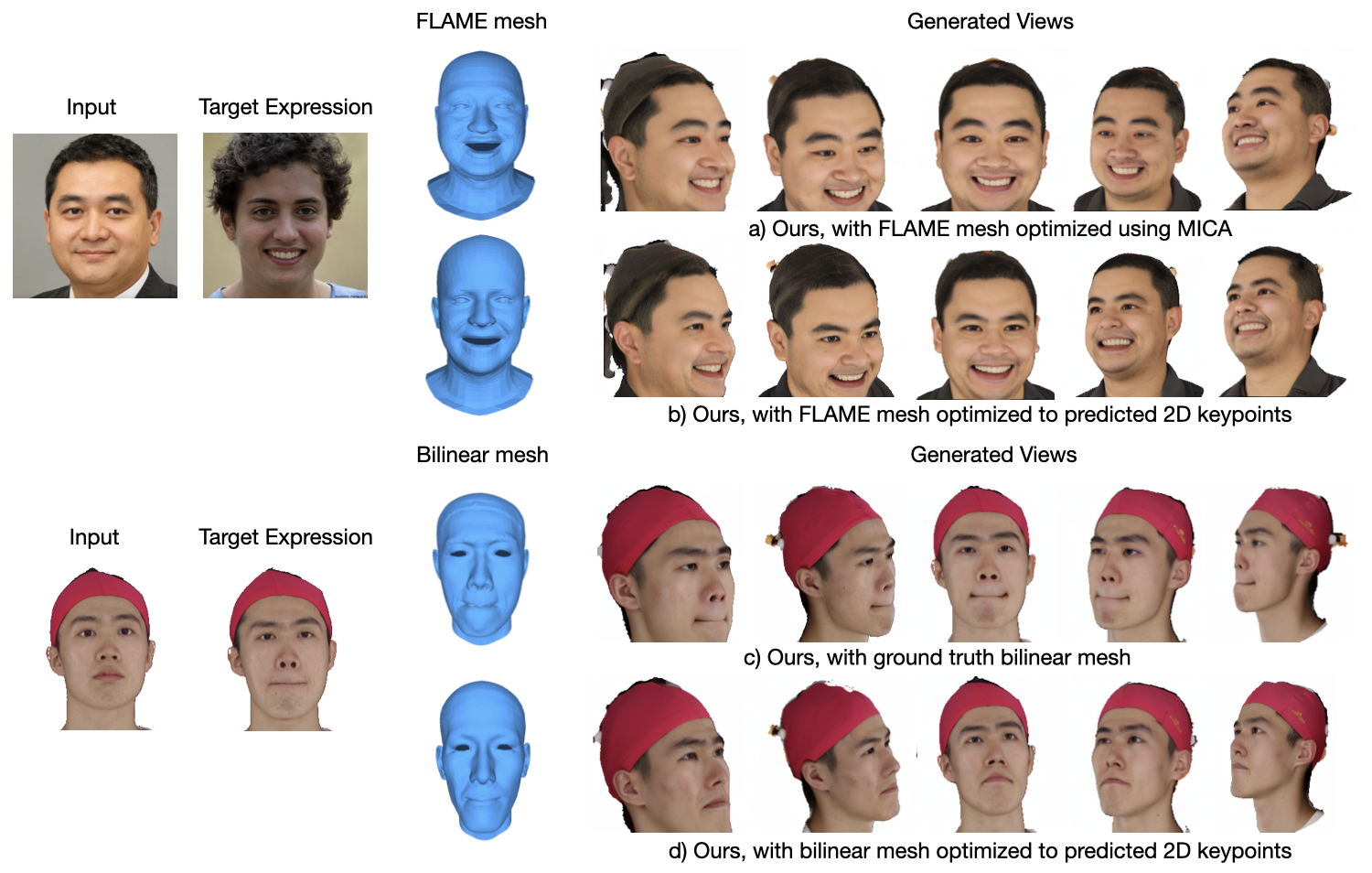}
\caption{\textbf{The effect of mesh accuracy on the generated faces}. We compare the generated faces using a FLAME mesh obtained with the state-of-the-art FLAME reconstruction pipeline \cite{MICA:ECCV2022} (a) with the generated results using the FLAME mesh fitted to detected 2D keypoints only via optimization (b). We also compare results using the ground truth bilinear mesh (c) with the results using the bilinear mesh fitted to 2D keypoints (d). (a) and (b) are tested using our model trained with FLAME models while (c) and (d) are tested using our model triained on bilinear models. Meshes with more accurately reconstructed facial expression and shape parameters lead to more accurate preservation of the target expression and better resemblance in the generated images.}
\label{fig:accuracy}
\end{figure}

Additionally, we find that having an accurate mesh estimator can lead to better preservation of the facial expression and better resemblance, as shown in Fig.~\ref{fig:accuracy}.

\end{document}